\pdfoutput=1

\documentclass[11pt]{article}

\usepackage[final]{EMNLP2022}

\usepackage{times}
\usepackage{latexsym}
\usepackage{graphicx}
\usepackage{subfig}
\usepackage{booktabs}

\usepackage[T1]{fontenc}

\usepackage[utf8]{inputenc}

\usepackage{microtype}

\usepackage{inconsolata}

\title{Large Language Models for Multi-label Propaganda Detection}

\author{Tanmay Chavan\thanks{~ Equal contribution}~~\and
    Aditya Kane$\footnotemark[1]$~ \\
  Pune Institute of Computer Technology, Pune \\
  \texttt{\{chavantanmay1402, adityakane1\}@gmail.com} }

\begin{document}
\maketitle
\begin{abstract}

The spread of propaganda through the internet has increased drastically over the past years. Lately, propaganda detection has started gaining importance because of the negative impact it has on society. In this work, we describe our approach for the WANLP 2022 shared task which handles the task of propaganda detection in a multi-label setting. The task demands the model to label the given text as having one or more types of propaganda techniques. There are a total of 21 propaganda techniques to be detected. We show that an ensemble of five models performs the best on the task, scoring a micro-F1 score of 59.73\%. We also conduct comprehensive ablations and propose various future directions for this work.

\end{abstract}

\section{Introduction}
The advent of social media has enabled people to view, create and share information easily on the internet. Such information can easily be accessed and viewed by a very large number of people in surprisingly short periods. Moreover, most social media websites have few restrictions over what the users choose to post and lack preemptive techniques to censor posts before they are uploaded. This has enabled the free flow of information from various strata of society which might have been restricted due to the lack of access to proper news sources. However, this has also led to a stark increase in the spread of propaganda through the internet. Information propagated through social media posts presents an individual's personal opinions, and hence is often biased and lacks rigorous fact-checking. Such problems are less frequently found in the original media sources of newspapers and TV news channels where their posts are subjected to a higher level of scrutiny. 

The presence of propaganda online poses a serious threat to society as it can often polarize the majority opinion and lead to violent events. A wave of misinformation-based propaganda during the time of the COVID-19 pandemic\cite{Cinelli_2020} was observed. However, the problem of propaganda detection is much more complicated than it appears. The biggest challenge in propaganda detection is that the bulk of propaganda information is partially based on truths, but is presented in a manner that might be misleading or unnecessarily polarizing. It is also observed that propaganda posts are written professionally and are compelling which makes most of the readers believe the information to be authentic. All of these problems make it difficult to train a model to detect propaganda, and much more difficult to interpret the results of such models. The purpose of the shared task \cite{propaganda-detection:WANLP2022-overview}, a multi-label classification problem, is to come up with efficient methods for detecting propaganda on a dataset containing Arabic tweets.

 Transformer-based models have achieved great success in text classification tasks. Additionally, ensemble-based models also outperform these individual models. Thus, we explore individual as well as ensemble of models for this task. Furthermore, we experimented with oversampling where we repeat the samples having minority labels. We also pretrained the DeHateBERT model on 1 million tweets to study the effect of domain-specific pretraining on downstream performance. We report the results of all these experiments and thereby propose an ensemble-based method for this task.

\begin{figure}[t]
    \centering
    \includegraphics[width=\columnwidth]{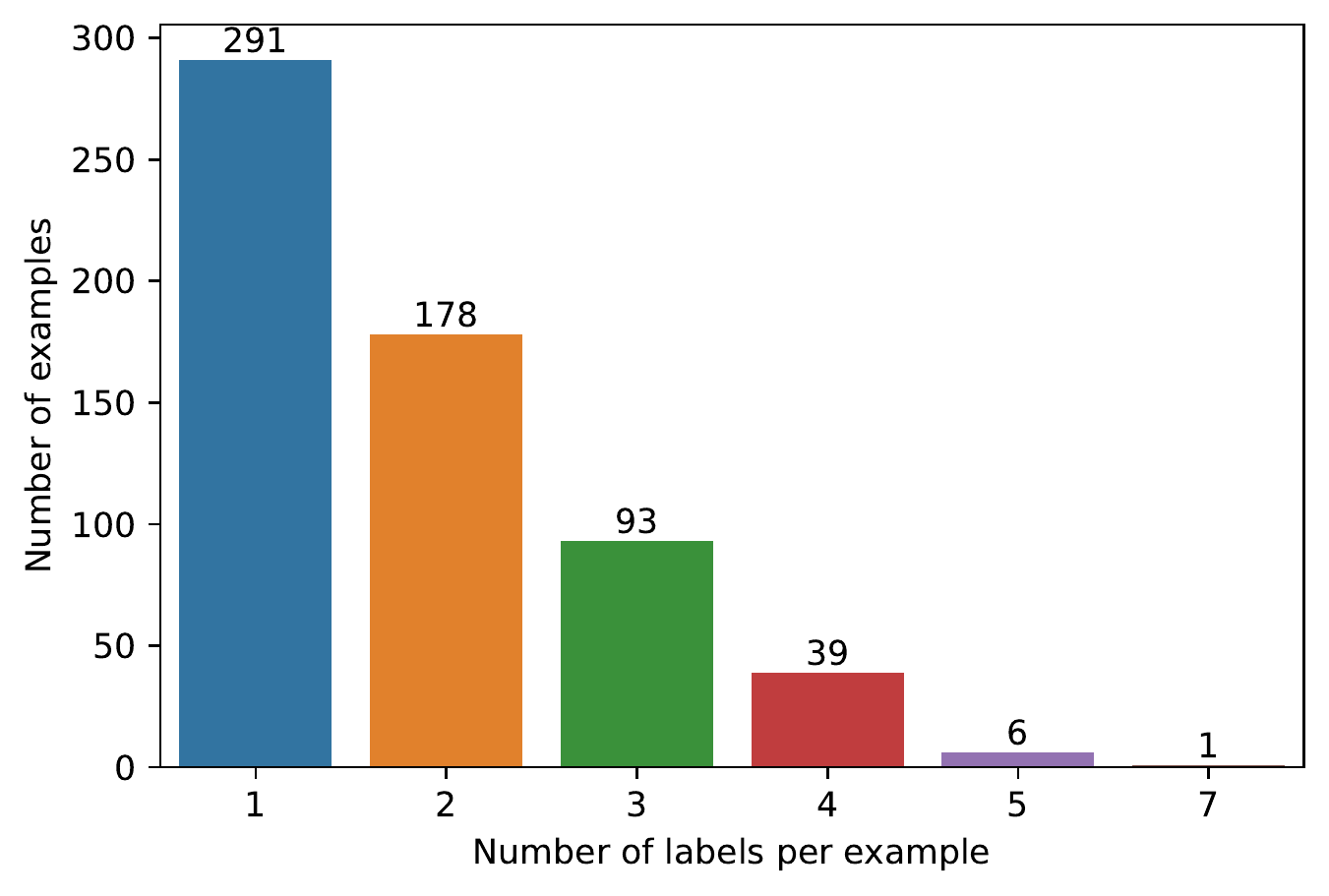}
    \caption{Distribution of label counts}
    \label{fig:label_counts}
\end{figure}

\begin{figure*}[t]
    \centering
    \includegraphics[width=\textwidth]{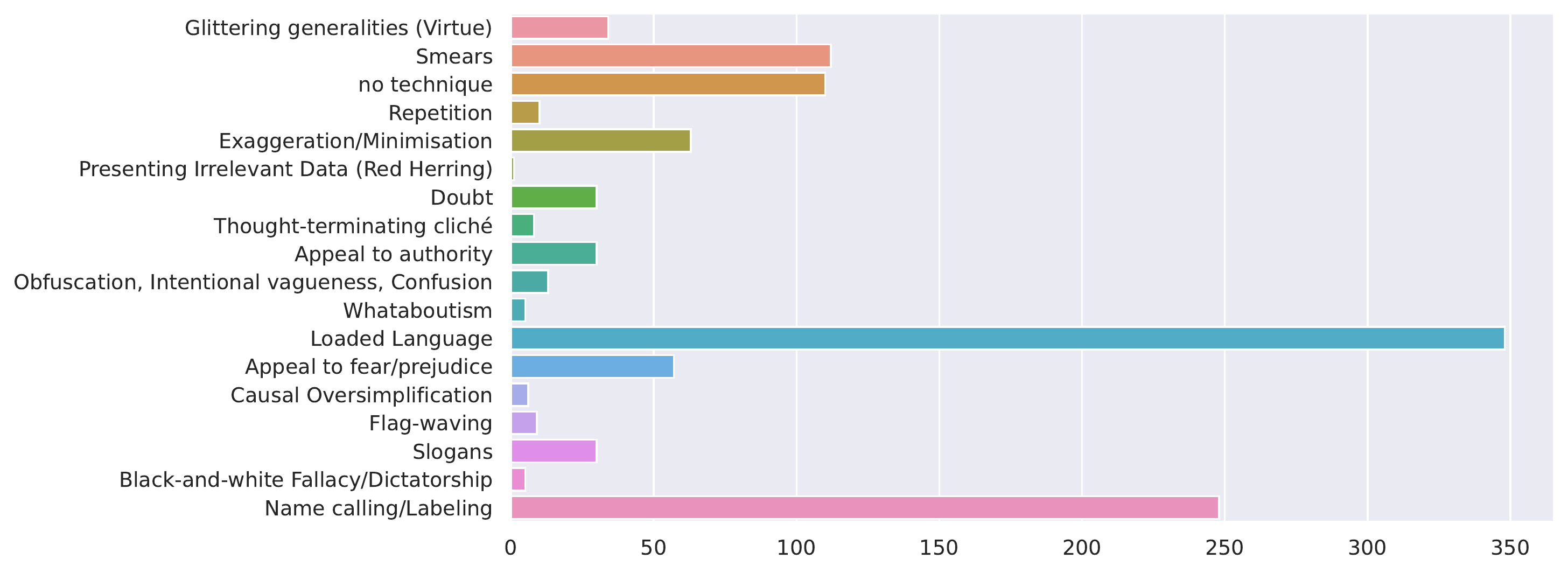}
    \caption{Data distribution}
    \label{fig:data_composition}
\end{figure*}

\section{Related Work}

\citet{da-san-martino-etal-2019-fine} effectively addressed the problem of quantifying different types of propaganda into seventeen categories, which helps us distinguish between different types of propaganda. They also presented a corpus that contains information classified according to the seventeen classes. Previous shared tasks have generated successful results. The SemEval 2020 task 11 \cite{da-san-martino-etal-2020-semeval} used the PTC corpus for building models to detect and classify propaganda. The SemEval task 6 \cite{dimitrov-etal-2021-semeval} helped develop novel approaches to detect propaganda in a multimodal environment. \citet{yu-etal-2021-interpretable} studied the topic of interpretability of propaganda detection and presented an interpretable model.

The use of BERT-based models which are pre-trained on a large corpus has proven to yield better performance than most of the other deep learning-based approaches without pre-training\cite{https://doi.org/10.48550/arxiv.2111.01243}. There are several BERT models pre-trained on massive Arabic datasets available. We test some of these models for the task. AraBERT \cite{antoun-etal-2020-arabert}, MARBERT, ARBERT \cite{abdul-mageed-etal-2021-arbert} are some examples. However, most of these models are pretrained on structured data which significantly differs from tweets. Research has shown that domain-specific pretraining can yield better performance than general text pretraining\cite{Brady2021AItBERTD}. Hence, we used DeHateBERT \cite{dehatebert}. 

\section{Data}

The dataset consists of 504 training examples, 52 validation examples, and 52 testing examples. Our models were finally evaluated on a separate testing dataset, which consisted of 323 examples. Each example can have one or more of the 20 propagandist techniques\footnote{Complete list of propagandist techniques can be found at \url{https://propaganda.qcri.org/annotations/definitions.html}}. Thus, it was a multi-label dataset. The number of label occurrences is illustrated in Figure \ref{fig:data_composition}. As shown in the figure, we see a skewed distribution. This shows that there is an imbalance. Given this problem of multi-label classification with a high class imbalance, we experimented with several architectures and found that DeHateBERT performed the best on the dataset. A full account of all of our successful experiments, as well as failed experiments, is given in Sections \ref{sec:results} and \ref{sec:discussion}.  We try multiple methods to mitigate this imbalance, as elaborated in Section \ref{sec:discussion}. Since the dataset is a multi-label dataset, used one-hot encoding for each label to denote the ground truth labels. 

Furthermore, we make some key observations about the number of labels per example in Figure \ref{fig:label_counts}. We observe that most examples have one label per example. We see that the number of examples having more than one label diminishes quickly, with only one example having 7 labels. 

We use basic preprocessing to minimize the noise in the inputs. Firstly, we remove all links in the tweet. Then we remove the user mentions and hashtags (denoted by "@" and "\#" followed by a string respectively). Finally, we replace underscores ("\_") with space. This way, the separated words contribute to the semantics of the sentence. Note that we retain the emojis in the sentence since they also carry significant meaning and can aid the model to better detect sentiment.

\section{System}

Given this problem of multi-label classification with a high class imbalance, we experimented with several architectures and found that DeHateBERT performed the best on the dataset. A full account of all of our successful experiments, as well as failed experiments, is given in Sections \ref{sec:results} and \ref{sec:discussion}. 


We tried several models, namely AraBERT v1, v02 and v2, MARBERT, ARBERT, XLMRoBERTa \cite{conneau-etal-2020-unsupervised}, AraELECTRA \cite{antoun-etal-2021-araelectra}. Note that the difference between AraBERTv2 and AraBERTv02 is that the former uses presegmented text whereas the latter uses the Farasa Segmenter \cite{darwish-mubarak-2016-farasa} to segment the text since Arabic is a language which requires its words to be segmented before being fed into the tokenizer. We used a specific variant of DeHateBERT, which is initialized from multilingual BERT and fine-tuned only on Arabic datasets. We found that this particular variant performed the amongst the best, in terms of micro-F1 on the test split of our dataset. Our model training is fairly straightforward. We train DeHateBERT on our multi-label dataset for 30 epochs and the best performing epoch is chosen based on validation micro-F1. We used a learning rate of $3e-6$. Note that we use binary cross-entropy loss, since we have multi-hot labels in our dataset. 

\begin{figure*}[t]
    \centering
    \includegraphics[width=\textwidth]{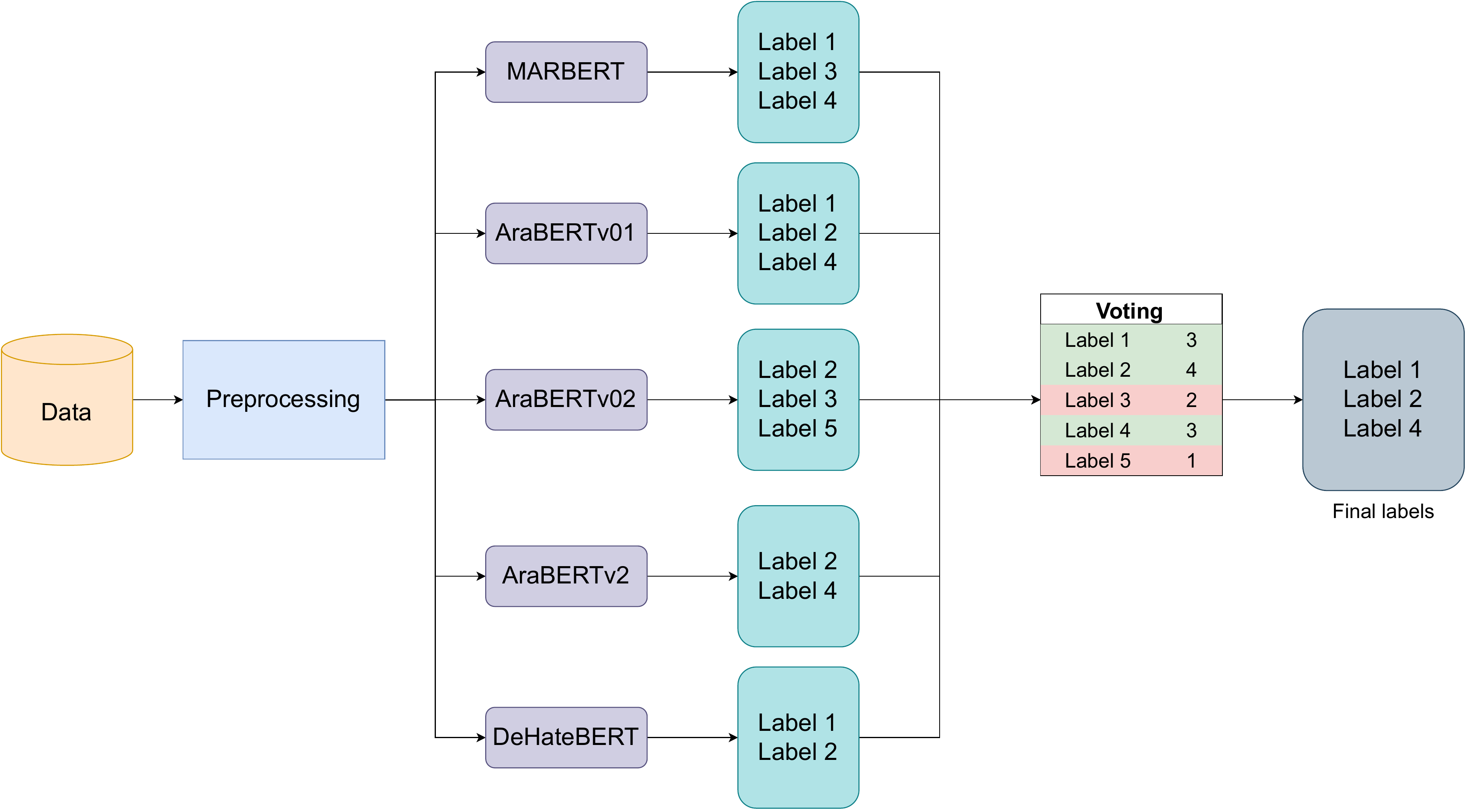}
    \caption{Ensemble system diagram. The ensemble works using a system of hard voting, wherein the prediction of each model is recorded and if the majority of models predict that label then it is declared to be one of the predicted labels. This figure illustrates this process in the multi-label setting.}
    \label{fig:ensemble}
\end{figure*}

We also create an ensemble of all the models. We use the five models namely DeHateBERT, ARBERT, AraBERTv02, AraBERTv01, and MARBERT. Our ensemble system is shown in Figure \ref{fig:ensemble}.  We use the method of hard voting to obtain the final results. For each sample, we recorded the predicted labels of each of the five models. Then, for each of the 21 labels present, we check how many models predict that label. If majority of the models predict the label, we include that label for the sample in the ensemble output. We find that the ensemble of models had the best performance. 

The dataset has a significant class imbalance. To overcome this, we tried to augment the dataset by oversampling. For oversampling, we duplicated the samples containing less frequent target classes. Thus we obtained a larger dataset containing duplicate samples but overall having lesser class imbalance. However, this did not yield better performance. We discuss this in detail in Section \ref{sec:discussion}.

\section{Results}
\label{sec:results}
The official scoring metric for the shared task is the F1 micro score. We present the results of the various models we tried in Table \ref{tab:final_results}. We have used the official scorer module provided by the organizers. We can see that the ensemble has the highest score. MARBERT and DeHateBERT have roughly similar scores and perform better than other models. This can lead us to speculate that a model might perform better at classification tasks if it is pretrained on a corpus containing data from a similar source than a corpus with similar characteristics but having data from a different source. The oversampled DeHateBERT model has a lower performance compared to the model trained on the original dataset.

We can however see that ARBERT outperforms all other single models. Another key observation is that ARBERT outperforms MARBERT, which in turn outperforms all variants of AraBERT. An explanation for this is that AraBERT variants are trained on far less data than ARBERT and MARBERT. In the case of ARBERT and MARBERT, ARBERT is pretrained on a wide variety of sources as opposed to MARBERT and thus has better performance than MARBERT.

We can also speculate that the high performance of the ensemble is because the constituent models are pretrained on different datasets. This enables the ensemble to capture a wider array of semantic vocabulary and hence is better at predicting classes. The hard voting mechanism ensures that the ensemble will not predict too many classes for each sample and thus limits the number of false positives.

\begin{table}[]
\begin{tabular}{|c|c|}
\hline
\textbf{Model}                     & \textbf{Micro-F1} \\ \hline
AraBERTv01                & 54.195   \\ \hline
AraBERTv2                 & 50.841   \\ \hline
AraBERTv02                & 53.996   \\ \hline
AraBERTv02-twitter        & 54.135   \\ \hline
DeHateBERT                & 56.484   \\ \hline
Oversampling + DeHateBERT & 52.529   \\ \hline
MARBERT                   & \textbf{56.556}  \\ \hline
ARBERT                    & \textbf{59.048}   \\ \hline
Ensemble                  & \textbf{59.725}   \\ \hline
\end{tabular}
\caption{Results of our experiments on the WANLP-22 propaganda detection task dataset.}
\label{tab:final_results}
\end{table}


\section{Discussion}
\label{sec:discussion}





We conducted several experiments apart from our best-performing model. Specifically, we tried pretraining on a large Arabic sentiment analysis tweet dataset as well as oversampling the classes having few samples. 

We retrained the DeHateBERT model on  1 Million tweets from the Large Arabic Twitter Data for Sentiment Analysis dataset using the Masked Language Modeling technique. We found that pretraining on the sentiment analysis tweet dataset did not result in any gains to the model. We speculate this is primarily because the number of tweets we pretrained the model on is less than the size the model was originally pretrained on. 

In another attempt, we implement oversampling in the dataset, where we repeat samples of less frequent classes. We calculate the average number of examples for each class. Then, we get the oversampling factor, that is the number of times the examples must be repeated to reach the average number of samples. We further clip this factor to 10. Note that, since this is a multi-label scenario, we need to be careful not to use examples with the most frequently occurring classes, in which case the process will have no effect.

Currently, we use hard voting for choosing the final output of the ensemble. We believe better results can be obtained by having a more sophisticated method like using an SVM instead of hard voting. 


\section{Conclusion}

This paper aims to articulate our approach for the WANLP 2022 Shared Task. We experimented with multiple transformer-based models, namely AraBERT, ARBERT, MARBERT and others. We also present ablations with monolingual pretraining, oversampling, and ensemble of the aforementioned transformer-based models. We show that the ensemble consisting of models pretrained on various sources of data has the best performance, with a Micro-F1 score of 59.73\%. We foresee several possible future directions. One line of work can be to improve the ensemble mechanism as well as to better handle the class imbalance in multi-label setting. Another line of work can be to study the effects of domain-specific pretraining on downstream classification tasks like multi-label classification.

\section*{Acknowledgement}

We thank Neeraja Kirtane for her reviews and inputs to this paper. 

\bibliography{anthology,custom}

\begin{thebibliography}{14}
\expandafter\ifx\csname natexlab\endcsname\relax\def\natexlab#1{#1}\fi

\bibitem[{Abdul-Mageed et~al.(2021)Abdul-Mageed, Elmadany, and
  Nagoudi}]{abdul-mageed-etal-2021-arbert}
Muhammad Abdul-Mageed, AbdelRahim Elmadany, and El~Moatez~Billah Nagoudi. 2021.
\newblock \href {https://doi.org/10.18653/v1/2021.acl-long.551} {{ARBERT} {\&}
  {MARBERT}: Deep bidirectional transformers for {A}rabic}.
\newblock In \emph{Proceedings of the 59th Annual Meeting of the Association
  for Computational Linguistics and the 11th International Joint Conference on
  Natural Language Processing (Volume 1: Long Papers)}, pages 7088--7105,
  Online. Association for Computational Linguistics.

\bibitem[{Alam et~al.(2022)Alam, Mubarak, Zaghouani, Nakov, and
  Da~San~Martino}]{propaganda-detection:WANLP2022-overview}
Firoj Alam, Hamdy Mubarak, Wajdi Zaghouani, Preslav Nakov, and Giovanni
  Da~San~Martino. 2022.
\newblock Overview of the {WANLP} 2022 shared task on propaganda detection in
  {A}rabic.
\newblock In \emph{Proceedings of the Seventh Arabic Natural Language
  Processing Workshop}, Abu Dhabi, UAE. Association for Computational
  Linguistics.

\bibitem[{Aluru et~al.(2020)Aluru, Mathew, Saha, and Mukherjee}]{dehatebert}
Sai~Saketh Aluru, Binny Mathew, Punyajoy Saha, and Animesh Mukherjee. 2020.
\newblock Deep learning models for multilingual hate speech detection.
\newblock \emph{CoRR}, abs/2004.06465.

\bibitem[{Antoun et~al.(2020)Antoun, Baly, and Hajj}]{antoun-etal-2020-arabert}
Wissam Antoun, Fady Baly, and Hazem Hajj. 2020.
\newblock \href {https://aclanthology.org/2020.osact-1.2} {{A}ra{BERT}:
  Transformer-based model for {A}rabic language understanding}.
\newblock In \emph{Proceedings of the 4th Workshop on Open-Source Arabic
  Corpora and Processing Tools, with a Shared Task on Offensive Language
  Detection}, pages 9--15, Marseille, France. European Language Resource
  Association.

\bibitem[{Antoun et~al.(2021)Antoun, Baly, and
  Hajj}]{antoun-etal-2021-araelectra}
Wissam Antoun, Fady Baly, and Hazem Hajj. 2021.
\newblock \href {https://aclanthology.org/2021.wanlp-1.20} {{A}ra{ELECTRA}:
  Pre-training text discriminators for {A}rabic language understanding}.
\newblock In \emph{Proceedings of the Sixth Arabic Natural Language Processing
  Workshop}, pages 191--195, Kyiv, Ukraine (Virtual). Association for
  Computational Linguistics.

\bibitem[{Brady(2021)}]{Brady2021AItBERTD}
Oliver~J. Brady. 2021.
\newblock Aitbert : Domain specific pretraining on alternative social media to
  improve hate speech classification.

\bibitem[{Cinelli et~al.(2020)Cinelli, Quattrociocchi, Galeazzi, Valensise,
  Brugnoli, Schmidt, Zola, Zollo, and Scala}]{Cinelli_2020}
Matteo Cinelli, Walter Quattrociocchi, Alessandro Galeazzi, Carlo~Michele
  Valensise, Emanuele Brugnoli, Ana~Lucia Schmidt, Paola Zola, Fabiana Zollo,
  and Antonio Scala. 2020.
\newblock \href {https://doi.org/10.1038/s41598-020-73510-5} {The {COVID}-19
  social media infodemic}.
\newblock \emph{Scientific Reports}, 10(1).

\bibitem[{Conneau et~al.(2020)Conneau, Khandelwal, Goyal, Chaudhary, Wenzek,
  Guzm{\'a}n, Grave, Ott, Zettlemoyer, and
  Stoyanov}]{conneau-etal-2020-unsupervised}
Alexis Conneau, Kartikay Khandelwal, Naman Goyal, Vishrav Chaudhary, Guillaume
  Wenzek, Francisco Guzm{\'a}n, Edouard Grave, Myle Ott, Luke Zettlemoyer, and
  Veselin Stoyanov. 2020.
\newblock \href {https://doi.org/10.18653/v1/2020.acl-main.747} {Unsupervised
  cross-lingual representation learning at scale}.
\newblock In \emph{Proceedings of the 58th Annual Meeting of the Association
  for Computational Linguistics}, pages 8440--8451, Online. Association for
  Computational Linguistics.

\bibitem[{Da~San~Martino et~al.(2020)Da~San~Martino, Barr{\'o}n-Cede{\~n}o,
  Wachsmuth, Petrov, and Nakov}]{da-san-martino-etal-2020-semeval}
Giovanni Da~San~Martino, Alberto Barr{\'o}n-Cede{\~n}o, Henning Wachsmuth,
  Rostislav Petrov, and Preslav Nakov. 2020.
\newblock \href {https://doi.org/10.18653/v1/2020.semeval-1.186}
  {{S}em{E}val-2020 task 11: Detection of propaganda techniques in news
  articles}.
\newblock In \emph{Proceedings of the Fourteenth Workshop on Semantic
  Evaluation}, pages 1377--1414, Barcelona (online). International Committee
  for Computational Linguistics.

\bibitem[{Da~San~Martino et~al.(2019)Da~San~Martino, Yu, Barr{\'o}n-Cede{\~n}o,
  Petrov, and Nakov}]{da-san-martino-etal-2019-fine}
Giovanni Da~San~Martino, Seunghak Yu, Alberto Barr{\'o}n-Cede{\~n}o, Rostislav
  Petrov, and Preslav Nakov. 2019.
\newblock \href {https://doi.org/10.18653/v1/D19-1565} {Fine-grained analysis
  of propaganda in news article}.
\newblock In \emph{Proceedings of the 2019 Conference on Empirical Methods in
  Natural Language Processing and the 9th International Joint Conference on
  Natural Language Processing (EMNLP-IJCNLP)}, pages 5636--5646, Hong Kong,
  China. Association for Computational Linguistics.

\bibitem[{Darwish and Mubarak(2016)}]{darwish-mubarak-2016-farasa}
Kareem Darwish and Hamdy Mubarak. 2016.
\newblock \href {https://aclanthology.org/L16-1170} {{F}arasa: A new fast and
  accurate {A}rabic word segmenter}.
\newblock In \emph{Proceedings of the Tenth International Conference on
  Language Resources and Evaluation ({LREC}'16)}, pages 1070--1074,
  Portoro{\v{z}}, Slovenia. European Language Resources Association (ELRA).

\bibitem[{Dimitrov et~al.(2021)Dimitrov, Bin~Ali, Shaar, Alam, Silvestri,
  Firooz, Nakov, and Da~San~Martino}]{dimitrov-etal-2021-semeval}
Dimitar Dimitrov, Bishr Bin~Ali, Shaden Shaar, Firoj Alam, Fabrizio Silvestri,
  Hamed Firooz, Preslav Nakov, and Giovanni Da~San~Martino. 2021.
\newblock \href {https://doi.org/10.18653/v1/2021.semeval-1.7}
  {{S}em{E}val-2021 task 6: Detection of persuasion techniques in texts and
  images}.
\newblock In \emph{Proceedings of the 15th International Workshop on Semantic
  Evaluation (SemEval-2021)}, pages 70--98, Online. Association for
  Computational Linguistics.

\bibitem[{Min et~al.(2021)Min, Ross, Sulem, Veyseh, Nguyen, Sainz, Agirre,
  Heinz, and Roth}]{https://doi.org/10.48550/arxiv.2111.01243}
Bonan Min, Hayley Ross, Elior Sulem, Amir Pouran~Ben Veyseh, Thien~Huu Nguyen,
  Oscar Sainz, Eneko Agirre, Ilana Heinz, and Dan Roth. 2021.
\newblock \href {https://doi.org/10.48550/ARXIV.2111.01243} {Recent advances in
  natural language processing via large pre-trained language models: A survey}.

\bibitem[{Yu et~al.(2021)Yu, Da~San~Martino, Mohtarami, Glass, and
  Nakov}]{yu-etal-2021-interpretable}
Seunghak Yu, Giovanni Da~San~Martino, Mitra Mohtarami, James Glass, and Preslav
  Nakov. 2021.
\newblock \href {https://aclanthology.org/2021.ranlp-1.179} {Interpretable
  propaganda detection in news articles}.
\newblock In \emph{Proceedings of the International Conference on Recent
  Advances in Natural Language Processing (RANLP 2021)}, pages 1597--1605, Held
  Online. INCOMA Ltd.

\end{thebibliography}
\bibliographystyle{acl_natbib}

\end{document}